\documentclass[letterpaper]{article}
\usepackage{uai2020}
\usepackage[margin=1in]{geometry}

\usepackage{times}
\usepackage[round]{natbib}

\usepackage[utf8]{inputenc}
\usepackage{url}
\RequirePackage[colorlinks=true,citecolor=blue,linkcolor=blue,urlcolor=blue]{hyperref}
\usepackage{amsfonts}
\usepackage{amsmath, amssymb, amsthm}
\usepackage{graphicx, xcolor}
\usepackage{booktabs}
\usepackage{framed}
\usepackage[nameinlink]{cleveref}
\usepackage{mathtools}
\usepackage{definitions}
\title{PAC-Bayesian Contrastive Unsupervised Representation Learning}
\author{ {\bf Kento Nozawa} \\
The University of Tokyo \& RIKEN \\
Japan \\
\And
{\bf Pascal Germain}  \\
Université Laval \\
Canada \\
\And
{\bf Benjamin Guedj}   \\
Inria \& University College London \\
France and United Kingdom \\
}

\begin{document}
\maketitle

\begin{abstract}
Contrastive unsupervised representation learning (CURL) is the state-of-the-art technique to learn representations (as a set of features) from unlabelled data.
While CURL has collected several empirical successes recently, theoretical understanding of its performance was still missing.
In a recent work, \citet{Arora2019ICML} provide the first generalisation bounds for CURL, relying on a Rademacher complexity.
We extend their framework to the flexible PAC-Bayes setting, allowing us to deal with the non-iid setting.
We present PAC-Bayesian generalisation bounds for CURL, which are then used to derive a new representation learning algorithm.
Numerical experiments on real-life datasets illustrate that our algorithm achieves competitive accuracy, and yields non-vacuous generalisation bounds.
\end{abstract}
\section{INTRODUCTION}
Unsupervised representation learning~\citep{Bengio2013IEEE} aims at extracting features representation from an unlabelled dataset for downstream tasks such as classification and clustering \citep[see][]{Mikolov2013NIPS, Noroozi2016ECCV, Zhang2016ECCV, Caron2018ECCV, Devlin2019NAACL}.
An unsupervised representation learning model is typically learnt by solving a pretext task without supervised information.
Trained model work as a feature extractor for supervised tasks.

In unsupervised representation learning, contrastive loss is a widely used objective function class.
Contrastive loss uses two types of data pair, namely, similar pair and dissimilar pair.
Their similarity is defined without label information of a supervised task.
For example, in word representation learning, \citet{Mikolov2013NIPS} define a similar pair as co-occurrence words in the same context,
while dissimilar pairs are randomly sampled from a fixed distribution.
Intuitively, by minimising a contrastive loss,
similar data samples are mapped to similar representations in feature space in terms of some underlying metric (as the inner product), and dissimilar samples are not mapped to similar representations.

Contrastive unsupervised representation learning improves the performance of supervised models in practice, and has attracted a lot of research interest lately \citep[see][and references therein]{Chen2020curl}, although usage is still quite far ahead of theoretical understanding.
Recently, \citet{Arora2019ICML} introduced a theoretical framework for contrastive unsupervised representation learning and derived the first generalisation bounds for CURL.
In parallel, PAC-Bayes is emerging as a principled tool to understand and quantify the generalisation ability of many machine learning algorithms, including deep neural networks \citep[as recently studied by][]{Dziugaite2017UAI,NeyshaburBS18,Letarte2019NeurIPS}.

\textbf{Our contributions.}
We extend the framework introduced by \citet{Arora2019ICML}, by adopting a PAC-Bayes approach to contrastive unsupervised representation learning.
We derive the first PAC-Bayes generalisation bounds for CURL, both in iid and non-iid settings.
Our bounds are then used to derive new CURL algorithms, for which we provide a complete implementation.
The paper closes with numerical experiments on two real-life datasets (\cifar{} and \auslan) showing that our bounds are non-vacuous in the iid setting.

\section{CONTRASTIVE UNSUPERVISED REPRESENTATION LEARNING}
\label{sec:curl}

\subsection{LEARNING FRAMEWORK}
Inputs are denoted $\xbf \in X = \Rbb^{d_0}$, and outputs are denoted $y \in Y$, where $Y$ is a discrete and finite set.

The \emph{representation} is learnt from a (large) unlabelled dataset $U=\{\zbf_i\}_{i=1}^{m}$, where $\zbf_i=(\xbf_i,\xbf_i^+, \xbf_{i1}^-, \ldots, \xbf_{ik}^-)$ is a tuple of $k{+}2$ elements; $\xbf_i$ being \emph{similar} to $\xbf_i^+$ and \emph{dissimilar} to every element of the \emph{negative sample set} $\{\xbf_{ij}^-\}_{j=1}^k$.
The predictor is learnt from a labelled dataset~$S= \{(\xbf_i, y_i)\}_{i=1}^{n}$.

In the following, we present the contrastive framework proposed by \citet{Arora2019ICML} in a simplified scenario in order to highlight the key ideas, 
where the supervised prediction task is binary and the negative sample sets for unsupervised representation learning contain one element.
Thus, we choose the label set to be $Y=\{-1,1\}$, and the unsupervised set $U$ contains triplets $\zbf_i=(\xbf_i,\xbf_i^+, \xbf_{i}^-)$.
The extension to a more generic setting (for $|Y|>2$ and $k>1$) bears no particular difficulty and is deferred to~\cref{sec:k-neg-bound}.
It is important to note at this stage that both $U$ and $S$ are assumed to be \textbf{iid} (independent, identically distributed) collections, as also assumed by \citet{Arora2019ICML}.

\textbf{Latent classes and data distributions.}
The main assumption is the existence of a set of \emph{latent classes} $\Ccal$. Let us denote by $\rho$ a probability distribution over $\Ccal$.
Moreover, with each class $c \in \Ccal$, comes a class distribution~$\Dcal_c$ over the input space $X$.
A similar pair $(\xbf, \xbf^+)$ is such that both $\xbf$ and $\xbf^+$ are generated by the same class distribution.
Note that an input $\xbf$ possibly belongs to multiple classes: take the example of $\xbf$ being an image and $\Ccal$ a set of latent classes including ``the image depicts a dog" and ``the image depicts a cat" (both classes are not mutually exclusive).
\begin{definition} \rm
    Let $\rho^2$ be a shorthand for the joint distribution $(\rho,\rho)$. We refer to the \emph{unsupervised data distribution} $\Ucal$ as the process that generates an unlabelled sample $\zbf=(\xbf,\xbf^+, \xbf^-)$ according to the following scheme:\\
        1. Draw two latent classes $(c^+, c^-) \sim \rho^2$\,;\\
        2. Draw two similar samples $(\xbf, \xbf^+) \sim (\Dcal_{c^{+}})^2$\,;\\
        3. Draw a negative sample $\xbf^- \sim \Dcal_{c^-}$\,.
\end{definition}
The labelled sample $S$ is obtained by fixing two classes $\cpm = \{c^-, c^+\} \in \Ccal^2$ (from now on, the shorthand notation  $\cpm$ is used to refer to a pair of latent classes).
Each class is then mapped on a label of $Y$.
We fix $y_{c^-} = -1$ and $y_{c^+} = 1$;
Thus we can write $Y=\{y_{c^-}, y_{c^+}\}$ as an ordered set.
The label is obtained from the latent class distribution restricted to two values $\rho_{\cpm}$:
$$\rho_{\cpm}(c^-) = \frac{\rho(c^-)}{\rho(c^-)\!+\!\rho(c^+)},
\ \rho_{\cpm}(c^+) = \frac{\rho(c^+)}{\rho(c^-)\!+\!\rho(c^+)}\,.$$
\begin{definition} \rm
    We refer to the \emph{supervised data distribution} $\Scal$ as the process that generates a labelled sample $(\xbf, y)$ according to the following scheme:\\
        1. Draw a class $c \sim \rho_{\cpm}$ and set label $y = y_{c}$\,;\\
        2. Draw a sample $\xbf\sim \Dcal_{c}$\,.
\end{definition}

\textbf{Loss function.} The learning process is divided in two sequential steps, the unsupervised and supervised steps.
In order to relate these two steps, the key is to express them in terms of a common convex loss function $\ell:\Rbb{\to}\Rbb^+$.
Typical choices are
\begin{align}
\label{eq:logloss}
    \ell_{\rm log} (v) \eqdef& \log_2(1+e^{-v})\,, & \mbox{\it(logistic loss)}\\
\label{eq:hingeloss}
    \ell_{\rm hinge} (v) \eqdef& \max\{0, 1{-}v\}\,, & \mbox{\it (hinge loss)}
\end{align}
where the loss argument $v$ expresses a notion of \emph{margin}.

In the first step, an unsupervised representation learning algorithm produces a feature map $\fbf : X \to \Rbb^d$.
The \emph{contrastive loss} associated with $\fbf$ is defined as
\begin{align*}
    \Lun(\fbf)  \eqdef \!\!\!& \Esp_{(c^+, c^-) \sim \rho^2} \, \Esp_{\substack{(\xbf, \xbf^+) \sim \Dcal_{c^+}^2\\ \xbf^-\sim\Dcal_{c^-}}}\!\!\!
    \ell \Big( \fbf(\xbf) {\cdot} \big[ \fbf(\xbf^+) {-} \fbf(\xbf^-)\big] \Big)\\
    =\!\!\!& \Esp_{(\xbf, \xbf^+, \xbf^-)\sim \Ucal} \ell \Big( \fbf(\xbf) {\cdot} \big[ \fbf(\xbf^+) {-} \fbf(\xbf^-)\big] \Big)\,.
\end{align*}
More precisely, from the unsupervised training dataset
\begin{equation}\label{eq:U}
U=\{(\xbf_i,\xbf_i^+, \xbf_{i}^-)\}_{i=1}^m\sim\Ucal^m\,,
\end{equation}
we are interested in learning the feature map $\fbf$ that minimises the following empirical contrastive loss:
\begin{align}
    \label{eq:Lunhatfbf}
    \Lunhat(\fbf) \eqdef \frac1m \sum_{i=1}^m \ell \Big( \fbf(\xbf_i) \cdot \big[ \fbf(\xbf_i^+) - \fbf(\xbf_i^-)\big] \Big)\,.
\end{align}
In the second step, a supervised learning algorithm is given the \emph{mapped} dataset
$\widehat{S} \,{\eqdef}\, \{(\hat \xbf_i, y_i)\}_{i=1}^{n}$,
with $\hat \xbf_i \,{\eqdef}\, \fbf(\xbf_i)$,
and returns a predictor $g:\Rbb^d \to \Rbb$.
For a fixed pair $\cpm = \{c^-,c^+\}$,
the predicted label on an input $\xbf$ is then obtained from $\hat y = \sgn[g(\hat \xbf)]$ (recall that $Y=\{-1,1\}$),
and we aim to minimise the supervised loss
\begin{align*}
    \Lsup(g\circ\fbf) \eqdef &
    \Esp_{c\sim \rho_\cpm} \Esp_{\xbf \sim \Dcal_c} \ell \Big( y_c \,g(\fbf(\xbf)) \Big) \\
    = &
    \Esp_{(\xbf, y) \sim \Scal} \ell \Big( y \,g(\fbf(\xbf)) \Big)\,.
\end{align*}
Given a labelled dataset $S\sim \Scal^n$, the empirical counterpart of the above supervised loss is
\begin{align*}
    \Lsuphat(g\circ\fbf) \eqdef \frac1n \sum_{i=1}^n \ell \Big( y_i \,g(\fbf(\xbf_i)) \Big)\,.
\end{align*}

\textbf{Mean classifier.}
Following \citet{Arora2019ICML}, we study the mean classifier defined by the linear function
\begin{align*}
    g_{\cpm}(\hat\xbf) \eqdef \wbf_{\cpm} \cdot \hat \xbf \,,
\end{align*}
where
$\wbf_{\cpm} \eqdef \mubf_{c^+} - \mubf_{c^-}$,
and $\mubf_{c} \eqdef \Esp_{\xbf \sim \Dcal_{c}} \fbf(\xbf)$.
Then, the \emph{supervised average loss} of the mean classifier is the expected loss on a dataset whose pair of labels is sampled from the latent class distribution $\rho$.
\begin{align} \label{eq:Lsupmuf}
    \Lsup^\mu(\fbf) \eqdef \Esp_{\cpm \sim \rhowo^2} \Lsup(g_{\cpm}\circ\fbf)\,,
\end{align}
with $\rhowo^2$ being a shorthand notation for the sampling \emph{without replacement} of two classes among $\Ccal$.
Indeed, we want positive and negative samples that are generated by distinct latent class distributions, \ie, $c^- \neq c^+$.
\subsection{GENERALISATION GUARANTEES}
A major contribution of the framework introduced by \citet{Arora2019ICML} is that it rigorously links the unsupervised representation learning task and the subsequent prediction task: it provides generalisation guarantees on the supervised average loss of \cref{eq:Lsupmuf} in terms of the empirical contrastive loss in \cref{eq:Lunhatfbf}.
Central to this result is the upcoming \cref{lemma:sup-bound}, that relates the supervised average loss of the mean classifier to its unsupervised loss.
\begin{lemma}[{\citealp[Lemma 4.3]{Arora2019ICML}}]
    \label{lemma:sup-bound}
    Given a latent class distribution $\rho$ on $\Ccal$ and a convex loss  $\ell:\Rbb{\to} \Rbb$, for any feature map $\fbf:\Rbb^{d_0}\to\Rbb^d$, we have
    \begin{align*}
        \Lsup^{\mu}(\fbf) \, \leq \,
        \frac{1}{1-\tau} \left( \Lun(\fbf) - \tau \right),
    \end{align*}
    where $\tau$ is the probability of sampling twice the same latent class
    ($\onebf[\cdot]$ is the indicator function):
    \begin{equation}
        \label{eq:tau}
                \tau \ \eqdef \Esp_{\cpm \sim \rho^2} \onebf[\cp = \cm]
                \,=\, \sum_{c\in\Ccal} \left[\rho(c)\right]^2.
    \end{equation}
\end{lemma}
\citet{Arora2019ICML} upper bound the unsupervised contrastive loss in \cref{lemma:sup-bound} by its empirical estimates.
The obtained generalisation guarantee is presented by the following \cref{theorem:Arora-bound}.
The bound focuses on a class of feature map functions $\Fcal$ through its empirical Rademacher complexity on a training dataset $U$, defined by
\begin{align*}
    \Rcal_U (\Fcal) \eqdef \Esp_{\sigmabf\sim\{\pm1\}^{3dm}} \left( \sup_{f\in\Fcal} \Big[  \sigmabf \cdot \fbf_{|U}\Big]\right),
\end{align*}
where $\fbf_{|U} \eqdef \vec(\{\fbf(\xbf_i), \fbf(\xbf^+_i), \fbf(\xbf^-_i)\}_{i=1}^m) \in\Rbb^{3dm}$ is the concatenation of all feature mapping given by $\fbf$ on $U$, and $\sigmabf{\sim}\{\pm1\}^{3dm}$ denotes the uniformly sampled Rademacher variables over that ``representation'' space.
\begin{theorem}[{\citealp[Theorem 4.1]{Arora2019ICML}}]
    \label{theorem:Arora-bound}
    Let \text{$B \in \Rbb_+$} be such that $\|\fbf(\cdot)\| \leq B$,
    with probability $1-\delta$ over training samples $U \sim \Ucal^m$, $\forall \fbf {\in} \Fcal$
    \begin{align*}
        &\Lsup^{\mu}(\widehat{\fbf}) \leq \\
        & \ \frac{1}{1 {-} \tau} \left( \Lun(\fbf) - \tau \right) + \frac{1}{1 {-} \tau}\, \mathcal{O}\!\left(B \tfrac{\Rcal_U(\Fcal)}{m} + B^2 \sqrt{\tfrac{\ln \frac{1}{\delta}}{m}} \right)\!,
    \end{align*}
    where $\hat \fbf \eqdef \displaystyle\argmin_{\fbf\in\Fcal} \Lunhat(\fbf)$\,.
\end{theorem}
\section{PAC-BAYES ANALYSIS}
Among the different techniques to analyse generalisation in statistical learning theory, PAC-Bayes has emerged in the late 90s as a promising alternative to the Rademacher complexity.
PAC-Bayes (\citealp[pioneered by][]{ShaweTaylor1997COLT,McAllester1998COLT,catoni2003pac,catoni2004statistical,Catoni2007} -- see \citealp{Guedj2019arXiv} for a recent survey) consists in obtaining PAC (probably approximately correct, \citealp{valiant1984theory}) generalisation bounds for Bayesian-flavoured predictors.
PAC-Bayes bounds typically hold with arbitrarily high probability and express a trade-off between the empirical risk on the training set and a measure of complexity of the predictors class.
A particularity of PAC-Bayes is that the complexity term relies on a divergence measure between a prior belief and a data-dependent posterior distribution (typically the Kullback-Leibler divergence).
\subsection{SUPERVISED LEARNING FRAMEWORK}
Let $\Pcal$ be a prior over a predictor class $\Hcal$, which cannot depend on training data,
and let $\Qcal$ be a posterior over the predictor class $\Hcal$, which can depend on the training data.
Any predictor $h\in\Hcal$ is a classification function $h:X \to Y$.
Most PAC-Bayes results measure the \emph{discrepancy} between the prior and the posterior distributions through the Kullback-Leibler divergence,
\begin{align}
    \label{eq:kl-divergence}
    \KL(\Pcal  \| \Qcal) \eqdef
    \bbE_{h \sim \Pcal} \ln\frac{\Pcal(h)}{\Qcal(h)}\,.
\end{align}
Moreover, PAC-Bayes provides bounds on the expected loss of the predictors under the distribution~$\Qcal$.
Let us present the classical supervised setup, where the zero-one loss is used.\footnote{Classical PAC-Bayes analyses consider the supervised learning setting, but non-supervised learning approaches exist~\citep[\eg,][]{Seldin2010JMLR,HiggsS10,Germain2013ICML}.}
We refer to this loss as the classification risk, denoted by $r(y,\hat y)\eqdef \onebf[y\, \hat y < 0]$.\footnote{See \cref{sec:contrastive-k-zero-one-risk} for a contrastive risk with $k$ negative samples.}
Given a data-generating distribution $\Scal$ on $X\times Y$, the expected $\Qcal$-risk is
\begin{align*}
    R(\Qcal) \eqdef \bbE_{(\xbf, y) \sim \Scal}  \bbE_{h \sim \Qcal} r(y,h(\xbf))\,,
\end{align*}
and the empirical counterpart, \ie{}, the $\Qcal$-weighted empirical risk on a training set $S=\{(\xbf_i, y_i)\}_{i=1}^n \sim \Scal^n$,  is given by
\begin{align*}
    \Rhat(\Qcal) \eqdef \frac{1}{n} \sum_{i=1}^n \bbE_{h \sim \Qcal} r(y_i, h(\xbf_i))\,.
\end{align*}
The following \cref{theorem:catoni-pac-bayes-bound} expresses an upper bound on the risk $R(\Qcal)$, from the empirical risk $\Rhat(\Qcal)$ and the posterior-prior divergence $\KL(\Qcal  \| \Pcal)$.
\begin{theorem}[{\citealp[Theorem 1.2.6]{Catoni2007}}]
    \label{theorem:catoni-pac-bayes-bound}
    Given $\lambda > 0$ and a prior $\Pcal$ over $\Hcal$,
    with probability at least $1-\delta$ over training samples $S\sim \Scal^n$, $\forall \Qcal$ over $\Hcal$,
    \begin{align}
        R(\Qcal) \leq
            \frac{
                1 - \exp \left(
                    - \lambda \Rhat(\Qcal)
                    - \frac{\KL(\Qcal \| \Pcal) + \ln\frac{1}{\delta}}{n}
                \right)
            }{
                1 - \exp \left(-\lambda \right)
            }\,.
    \end{align}
\end{theorem}

\subsection{PAC-BAYES REPRESENTATION LEARNING}
We now proceed to the first of our contributions.
We prove a PAC-Bayesian bound on the contrastive unsupervised representation loss, by replacing the Rademacher complexity in ~\cref{theorem:Arora-bound} with a Kullback-Leibler divergence.
To do so, we consider a prior $\Pcal$ and posterior $\Qcal$ distributions over a class of feature mapping functions $\Fcal \eqdef \{\fbf\,{\in}\,X{\to} \Rbb^d\}$.
Note that our PAC-Bayesian analysis for a multi-class extension is found at~\cref{sec:k-neg-bound}.

First, let us remark that we can adapt~\cref{theorem:catoni-pac-bayes-bound} to a bound on the unsupervised expected \emph{contrastive risk} defined as
\begin{align*}
    \Run(\Qcal) \eqdef \Esp_{(\xbf, \xbf^+, \xbf^-)\sim \Ucal} \Esp_{\fbf\sim \Qcal}
    r \Big( \fbf(\xbf^+) {-} \fbf(\xbf^-), \fbf(\xbf) \Big),
\end{align*}
where $r(\ybf, \hat\ybf) \eqdef \onebf[\ybf\cdot\hat\ybf<0]$ is the zero-one loss extended to vector arguments.
We denote $\Runhat(\Qcal)$ the empirical counterpart of $\Run(\Qcal)$ computed on the unsupervised training set $U\sim \Ucal^m$.
Once expressed this way, \cref{theorem:catoni-pac-bayes-bound}---devoted to classical supervised learning---can be straightforwardly adapted for the expected contrastive risk.
Thus, we obtain the following \cref{corollary:catoni-unsup-bound}.
\begin{corollary}
    \label{corollary:catoni-unsup-bound}
    Given $\lambda > 0$ and a prior $\Pcal$ over $\Fcal$,
    with probability at least $1-\delta$ over training samples $U\sim\Ucal^m$, $\forall \Qcal$ over $\Fcal$,
    \begin{align*}
        \Run(\Qcal) \leq
            \frac{
                1 - \exp \left(
                    - \lambda \Runhat(\Qcal)
                    - \frac{\KL(\Qcal \| \Pcal) + \ln\frac{1}{\delta}}{m}
                \right)
            }{
                1 - \exp \left(-\lambda \right)
            }\,.
    \end{align*}
\end{corollary}
Unfortunately, the bound on the contrastive risk $\Run(\cdot)$ does not translate directly to a bound on the supervised average risk
\begin{align} \label{eq:Rsupmuf}
    \Rsup^\mu(\fbf) \eqdef \Esp_{\cpm \sim \rhowo^2} \Rsup(g_{\cpm}\circ\fbf)\,.
\end{align}
This is because the zero-one loss is not convex, preventing us from applying \cref{lemma:sup-bound} to obtain a result analogous to \cref{theorem:Arora-bound}.
However, note that both loss functions defined by Equations~(\mbox{\ref{eq:logloss}-\ref{eq:hingeloss}}) are upper bound on the zero-one loss:
\begin{align*}
    \forall \ybf, \hat \ybf \in \Rbb^d : r(\ybf, \hat \ybf) \leq \ell(\ybf\cdot\hat\ybf)\,,
    \mbox{ with } \ell\in\{\ell_{\rm log}, \ell_{\rm hinge}\}.
\end{align*}
Henceforth, we study the $\Qcal$ expected loss
$$\Lsup^{\mu}(\Qcal) = \bbE_{\fbf \sim \Qcal} \Lsup^{\mu}(\fbf )$$ in regards to
$$\Lun(\Qcal) = \bbE_{\fbf  \sim \Qcal} \Lun(\fbf )\,.$$
By assuming that the representation vectors are bounded, \ie, $\|\fbf(\cdot)\| \leq B$ for some $B\in\Rbb^+$ as in \cref{theorem:Arora-bound}, we also have that the loss function is bounded.
Thus, by rescaling in $[0,1]$ the loss function, \cref{theorem:catoni-pac-bayes-bound} can be used to derive
the following \cref{theorem:iid-PAC-Bayesian-contrastive-bound}, which is the PAC-Bayesian doppelg\"anger of \cref{theorem:Arora-bound}.
\begin{theorem}
    \label{theorem:iid-PAC-Bayesian-contrastive-bound}
    Let $B \in \Rbb_+$ such that $\|\fbf(\cdot)\| \leq B$ for all $\fbf\in\Fcal$.
    Given $\lambda > 0$ and a prior $\Pcal$ over $\Fcal$,
    with probability at least $1-\delta$ over training samples $U\sim\Ucal^m$, $\forall \Qcal$ over $\Fcal$,
    \begin{align}
        \label{eq:pacbayesiid}
        &\Lsup^{\mu}(\Qcal) \leq \\
        &\frac{1}{1 {-} \tau}\!
        \Bigg(\!
        B_\ell\frac{
                1  {-} \exp \big(
                        {-} \frac{\lambda}{B_\ell} \Lunhat(\Qcal)
                        - \frac{\KL(\Qcal \| \Pcal) + \ln\frac{1}{\delta}}{m}
                    \big)
            }{
                1 - \exp(-\lambda)
            }
        - \tau \!\Bigg), \nonumber
    \end{align}
    with $B_\ell{\eqdef} \max\{\ell(-2B^2), \ell(2B^2)\}$
    and $\tau$ given by Eq.~\eqref{eq:tau}.
\end{theorem}
\begin{proof} Since $\|\fbf(\cdot)\| \leq B$, we have $\forall \xbf,\xbf^+,\xbf^-\in X^3$:
$$-2 B^2 \leq  \fbf(\xbf) \cdot [ \fbf(\xbf^+)-\fbf(\xbf^-) ] \leq 2 B^2\,. $$
Thus, $\ell(\fbf(\xbf)\cdot[\fbf(\xbf^+)-\fbf(\xbf^-)]) \leq B_\ell $, as $\ell$ is both convex and positive.
Therefore, the output of the rescaled loss function $\ell'(\cdot) \eqdef \frac{1}{B_\ell} \ell(\cdot)$  belongs to $[0,1]$.
From that point, we apply \cref{theorem:catoni-pac-bayes-bound} to obtain\footnote{ \cref{theorem:catoni-pac-bayes-bound} is given for the zero-one loss, but many works show that the same argument holds for any $[0,1]$-bounded loss \citep[\eg,][]{HiggsS10}.},
with probability at least $1-\delta$,
\begin{align*}
    \frac{1}{B_\ell} \Lun(\Qcal) \leq
        \frac{
            1 {-} \exp \Big(
                {-} \frac{\lambda}{B_\ell}\Lunhat(\Qcal)
                - \frac{\KL(\Qcal \| \Pcal) + \ln\frac{1}{\delta}}{m}
            \Big)
        }{
            1 - \exp \left(-\lambda \right)
        }.
\end{align*}
Also, since the inequality stated in \cref{lemma:sup-bound} holds true for all $\fbf \in \Fcal$, taking the expected value according to $\Qcal$ gives
    \begin{align*}
        \Lsup^{\mu}(\Qcal) \, \leq \,
        \frac{1}{1-\tau} \left( \Lun(\Qcal) - \tau \right).
    \end{align*}
The desired result is obtained by replacing $\Lun(\Qcal)$ in the equation above by its bound in terms of $\Lunhat(\Qcal)$.
\end{proof}
The Rademacher bound of \cref{theorem:Arora-bound} and the PAC-Bayes bound of \cref{theorem:iid-PAC-Bayesian-contrastive-bound} convey a similar message: finding a good representation mapping (in terms of the empirical contrastive loss) guarantee to generalise well, on average, on the supervised tasks.

An asset of the PAC-Bayesian bound lies in the fact that its exact value is easier to compute than the Rademacher one.
Indeed, for a well-chosen prior-posterior family, the complexity term $\KL(\Qcal\|\Pcal)$ has a closed-form solution, while computing $\Rcal_U(\Fcal)$ involves a combinatorial complexity.
From an algorithm design perspective, the fact that $\KL(\Qcal \| \Pcal)$ varies with $\Qcal$ suggests a trade-off between accuracy and complexity to drive the learning process,
while $\Rcal_U(\Fcal)$ is constant for a given choice of class~$\Fcal$.
We leverage these assets to propose a bound-driven optimisation procedure for neural networks in \cref{sec:algorithms}.

Note that one could be interested to study the risk of a predictor learned on the representation of the supervised data instead of the mean classifier's risk.
As discussed in~\cref{sec:general-analysis}, the loss of the best supervised predictor is at least as good as the mean classifier's one.
\subsection{RELAXING THE IID ASSUMPTION}
An interesting byproduct of \citet{Arora2019ICML}'s approach is that the proof of the main bound (\cref{theorem:Arora-bound}) is modular: we mean that in the proof of \cref{theorem:iid-PAC-Bayesian-contrastive-bound}, instead of plugging in Catoni's bound (\cref{theorem:catoni-pac-bayes-bound}), we can use any relevant bound.
We therefore leverage the recent work of \citet{Alquier2018ML} who proved a PAC-Bayes generalisation bound which no longer needs to assume that data are iid, and even holds when the data-generating distribution is heavy-tailed.
We can therefore cast our results onto the non-iid setting.

We believe removing the iid assumption is especially relevant for contrastive unsupervised learning, as we deal with triplets of data points governed by a relational causal link (similar and dissimilar examples).
In fact, several contrastive representation learning algorithms violate the iid assumption~\citep{Goroshin2015ICCV, Logeswaran2018ICLR}.

\pagebreak
\citet{Alquier2018ML}'s framework generalises the Kullback-Leibler divergence in the PAC-Bayes bound with the class of $f$-divergences \citep[see][for an introduction]{csiszar2004information}.
Given a convex function $f$ such that $f(1) = 0$,
the $f$-divergence between two probability distributions is given by
\begin{align}
    \label{eq:f-divergence}
    D_f(\Pcal \| \Qcal) = \bbE_{h \sim \Qcal} f \left( \frac{\Pcal(h)}{\Qcal(h)} \right).
\end{align}

\begin{theorem}
    \label{theorem:PAC-Bayes-contrastive}
    Given $p > 1, q = \frac{p}{p-1}$ and a prior $\Pcal$ over~$\Fcal$,
    with probability at least $1-\delta, \forall \Qcal$ over $\Fcal$,
    \begin{multline}
        \Lsup^{\mu}(\Qcal) \leq
        \frac{1}{1-\tau} \left( \Lunhat(\Qcal) - \tau \right) \\
        + \frac{1}{1 - \tau} \left( \frac{\Mcal_{q}}{\delta} \right)^{\frac{1}{q}}
        \left( D_{\phi_{p}-1}(\Qcal \| \Pcal) + 1 \right)^{\frac{1}{p}},
        \label{eq:pacbayes-noniid}
    \end{multline}
    where
    $\Mcal_{q} = \bbE_{\fbf \sim \Pcal} \bbE_{U \sim \Ucal^m}  (| \Lun(\fbf) - \Lunhat(\fbf) |^q )$ (recall that $\widehat{L}_\mathrm{un}$ depends on $U$, see Eqs.  \ref{eq:U} and \ref{eq:Lunhatfbf}) and $\phi_p(x) {=} x^p$.
\end{theorem}
The proof is a straightforward combination of aforementioned results, substituting Theorem 1 in~\citet{Alquier2018ML} to Catoni's bound (\cref{theorem:catoni-pac-bayes-bound}) in the proof of~\cref{theorem:iid-PAC-Bayesian-contrastive-bound}.
Up to our knowledge, \cref{theorem:PAC-Bayes-contrastive} is the first generalisation bound for contrastive unsupervised representation learning that holds without the iid assumption, therefore extending the framework introduced by~\citet{Arora2019ICML} in a non-trivial and promising direction.
Note that \cref{theorem:PAC-Bayes-contrastive} does not require iid assumption for both unsupervised and supervised steps.
\section{FROM BOUNDS TO ALGORITHMS}
\label{sec:algorithms}
In this section, we propose contrastive unsupervised representation learning algorithms derived from the PAC-Bayes bounds stated in \cref{theorem:iid-PAC-Bayesian-contrastive-bound,theorem:PAC-Bayes-contrastive}.
The algorithms are obtained by optimising the weights of a neural network by minimising the right-hand side of \eqref{eq:pacbayesiid} and \eqref{eq:pacbayes-noniid}, respectively.
Our training method is inspired by the work of \citet{Dziugaite2017UAI}, who optimise a PAC-Bayesian bound in a supervised classification framework,
and show that it leads to non-vacuous bounds values and accurately detects overfitting.
\subsection{NEURAL NETWORK OPTIMISATIONS}
\label{sec:NNO}
\subsubsection{Algorithm based on \cref{theorem:iid-PAC-Bayesian-contrastive-bound}}
\label{sec:iid-algorithm}
We consider a neural network architecture with $N$ real-valued learning parameters.
Let us denote $\wbf \in\Rbb^N$ the concatenation into a single vector of all the weights, and $\fbf_\wbf:X\to \Rbb^d$ the output of the neural network whose output is a $d$-dimensional \emph{representation} vector of its input.
From now on, $\Fcal_N=\{\fbf_\wbf|\wbf\in\Rbb^N \}$ is the set of all possible neural networks for the chosen architectures.
We restrict the posterior and prior over $\Fcal_N$ to be Gaussian distributions, that is
\begin{align*}
    \Qcal \eqdef \Ncal(\boldsymbol{\mu}_\Qcal, \diag(\sigmabf{}^2_{\Qcal}))\,, \quad
    \Pcal \eqdef \Ncal(\boldsymbol{\mu}_\Pcal, \sigma_{\Pcal}^2 I )\,,
\end{align*}
where $\boldsymbol{\mu}_\Qcal, \boldsymbol{\mu}_\Pcal \in \Rbb^{N}$,
$\sigmabf{}_{\Qcal}^2 \in \Rbb_+^N$,
and $\sigma_{\Pcal}^2 \in \Rbb_+$.

Given a fixed $\lambda$ in \cref{theorem:iid-PAC-Bayesian-contrastive-bound},
since $\tau$ is a constant value,
minimising the upper bound is equivalent to minimising the following expression\footnote{Note that without loss of generality, the constant $B_\ell$ is absorbed by $\lambda$ and plays no role in the optimisation objective.}
\begin{align}
    \lambda \,m\, \Lunhat(\Qcal)
    + \KL(\Qcal \| \Pcal) + \ln\frac{1}{\delta}\,.
\end{align}
Since $\Lunhat(\Qcal)$ is still intractable (as it is expressed as the expectation with respect to the posterior distribution on predictors),
we resort to an unbiased estimator;
the weight parameters are sampled at each iteration of a gradient descent, according to
$$\wbf = \boldsymbol{\mu}_\Qcal + \sigmabf{}_\Qcal \odot \boldsymbol{\epsilon}\,;
\mbox{ with }
\boldsymbol{\epsilon} \sim \Ncal(\mathbf{0}, I)\,,$$
the symbol $\odot$ being the element-wise product.
Therefore we optimise the posterior's parameters $\boldsymbol{\mu}_{\Qcal}$ and $\sigmabf{}_{\Qcal}^2$.
In addition, we optimise the prior variance $\sigma^2_{\Pcal}$ in the same way as~\citet[Section~3.1]{Dziugaite2017UAI}.
That is, given fixed $b, c\in\mathbb{R}_+$, we consider the bound value for
\begin{align} \label{eq:j}
    \sigma_\Pcal^2 \in \{ c \exp\big({-}\tfrac j b\big)\mid j \in \mathbb{N} \}\,.
\end{align}
From the union bound argument, the obtained result is valid with probability $1-\delta$ by computing each bound with a confidence parameter $\delta_j \eqdef 1-\frac{6}{\pi^2j^2}$, where $j=b \ln\tfrac{c}{\sigma^2_{\Pcal}}$.

Given $\delta, b, c,$ and $\lambda$, our final objective based on \cref{theorem:iid-PAC-Bayesian-contrastive-bound} is
\begin{align*}
    \min_{\boldsymbol{\mu}_{\Qcal, \boldsymbol{\sigma^2_{\Qcal}}, \sigma^2_{\Pcal}}}
    \hspace{-3mm} \lambda \,m\,\Lunhat(\Qcal)
    +  \KL(\Qcal \| \Pcal)
        + 2 \ln \big( b \ln\tfrac{c}{\sigma^2_{\Pcal}} \big)\,,
\end{align*}
where
\begin{align*}
&\KL(\Qcal \| \Pcal ) = \\[-1mm]
&\ \ \frac{1}{2} \!\Big(
    \tfrac{\| \boldsymbol{\mu}_\Qcal - \boldsymbol{\mu}_\Pcal \|_2^2}{\sigma^2_{\Pcal}}
    \!-\! N \!+\! \tfrac{ \| \sigmabf{}^2_{\Qcal} \|_1 }{\sigma^2_\Pcal}
    \!+\! N \ln \sigma^2_{\Pcal}
    \!-\! \sum_{i=1}^N
    \ln \sigmabf{}^2_{\Qcal, i}
\Big)\,.
\end{align*}
\subsubsection{Algorithm based on \cref{theorem:PAC-Bayes-contrastive}}
\label{sec:non-iid-algorithm}
We consider the same neural network architecture, prior, and posterior as in \cref{sec:iid-algorithm}.

We specify $p=2$ in \cref{theorem:PAC-Bayes-contrastive} to use a familiar $f$-divergence: the $\chi^2$-divergence.
Then, minimising the upper bound is equivalent to minimising the following expression:
\begin{align}
    \Lunhat(\Qcal) +
    \sqrt{ \frac{\Mcal_2}{\delta} \left( \chi^2(\Qcal \| \Pcal) +1 \right) }.
    \label{eq:general-objective}
\end{align}
Even though we use the unbiased estimator to evaluate the first term like iid algorithm,
the objective is still intractable since the moment $\Mcal_{2}$ requires the test loss $\Lun(\fbf)$.
Thus we assume the existence of an upper bound on the covariance of the contrastive loss $\ell$ to bound $\Mcal_2$ as follow\footnote{More generally, we may use $\alpha$-mixing based upper bound of the moment described by~\citet{Alquier2018ML}.}:
\begin{align}
    \mathrm{Cov}(\ell(\zbf_i), \ell(\zbf_j)) \begin{cases}
        \leq B_{\ell}^2 & \mbox{if } i {-} T \leq j \leq i {+} T \\
        = 0 &  \mbox{otherwise}
    \end{cases},
    \label{eq:upper-bound-cov}
\end{align}
where $T$ is the length of dependency to generate similarity pairs $(\xbf, \xbf^+)$.

This assumption is natural for CURL on sequential data~\citep{Mikolov2013NIPS, Goroshin2015ICCV},
where a positive sample $\xbf^+$ appears in sample $\xbf$'s neighbours in a time series.

Given $\delta, b, c,$ and $T$,
our final objective is
\begin{align} \nonumber
    &\min_{\mubf_\Qcal, \sigmabf{}^2_{\Qcal}, \sigma^2_{\Pcal}}
    \Lunhat(\Qcal)
    +{}
    \\
    &\pi \left(b \ln \frac{c}{\sigma_\Pcal^2} \right)
    \sqrt{
        \frac{ B_{\ell}^2 }{24m \delta} ( 1 + 8T )
        \left(
            \chi^2(\Qcal \| \Pcal) + 1
        \right)
    },
    \label{eq:non-iid-objective}
\end{align}
where the full expression of $\chi^2$-divergence is found in \cref{sec:full-chi-square-divergence}.
The objective is obtained by using the covariance's assumption and the union bound for the prior's variance $\sigma_\Pcal^2$.

The objective value is large if $T$ is large, that is when data dependency is long.
Therefore collecting independent time-series samples is a more effective way to tighten the bound than increasing $T$.
Interestingly, \cref{eq:general-objective} with \cref{eq:upper-bound-cov} can be viewed as a generalised bound of~\citet[Corollary 10]{Begin2016AISTATS}.
In fact, our objective becomes their bound when the data is iid and $\ell$ is the zero-one loss.
\subsection{PARAMETER SELECTION}
\label{sec:parameter-selection}
In the forthcoming experiments (Section~\ref{section:experiments}), we empirically compare the following three criteria for parameter selection: (i) the validation contrastive risk according to the posterior $\Qcal$, (ii) the validation contrastive risk of the \emph{maximum a posteriori network}, and (iii) the PAC-Bayes bound associated with the learned $\Qcal$.

For the first validation contrastive risk criterion,
we select a model with the best hyper-parameters such that it achieves the lowest contrastive risk $\Lunhat(\Qcal)$ on the validation data.
We approximate $\Lunhat(\Qcal)$ in a Monte Carlo fashion by sampling several $\fbf_\wbf$ from $\Qcal$.

Empirically, stochastic neural networks learnt by minimising the PAC-Bayes bound perform quite conservatively~\citep{Dziugaite2017UAI}.
Therefore we also use a validation contrastive risk computed with the deterministic neural network being the most likely according to the posterior (\ie, the neural network weights are taken as the mean vector of the posterior, rather than sampled from it).

The last criterion, the PAC-Bayes bound, does not use validation data; it only requires training data.
For the algorithm described in \cref{sec:iid-algorithm},
we select a model with the best hyper-parameters such that it minimises the following PAC-Bayes bound on the contrastive supervised risk $\Run(\Qcal)$:
\begin{align}
    \label{eq:parameter-selection-criterion}
    \min_{\lambda > 0}\!\Bigg[
    \tfrac{
            1{-}\exp \Big(
                    - \lambda \Runhat(\Qcal)
                    - \tfrac{\KL(\Qcal \| \Pcal) + \ln \frac{\pi^2 j^2}{6} + \ln \frac{2\sqrt{m} }{\delta}}{m}
                \Big)
        }{
            1 - \exp({-\lambda})
        }\Bigg]\!.
\end{align}
This criterion is given by \cref{corollary:catoni-unsup-bound},
where the term $\ln\frac1\delta$ is replaced by $\ln \frac{\pi^2 j^2}{6} + \ln \frac{2\sqrt{m} }{\delta}$.
The first summand comes from the union bound over the prior's variances--see \cref{eq:j}.
The second summand replaces $\frac1\delta$ by $\frac{2\sqrt{m}}{\delta}$,
as~\citet[Theorem 3]{Letarte2019NeurIPS} showed that this suffices to make the bound valid uniformly for all $\lambda>0$,
which allows for minimising the bound over $\lambda$.
Note that the learning algorithm minimises a bound on the (differentiable) convex loss,
but our model selection bound focuses on the zero-one loss as our task is a classification one.
\section{NUMERICAL EXPERIMENTS}
\label{section:experiments}
Our experimental codes are publicly available.
\footnote{\url{https://github.com/nzw0301/pb-contrastive}} 
We implemented all algorithms with \texttt{PyTorch}~\citep{Paszke2019NeurIPS}.
Herein, we report experiments for the algorithm described in \cref{sec:iid-algorithm}.
Experiments for the non-iid algorithm are provided in \cref{sec:non-iid-experiments}.
\subsection{PROTOCOL}
\textbf{Datasets.}
We use \cifar~\citep{Krizhevsky2009techrep} image classification task,
containing $60\,000$ images, equally distributed into $100$ labels.
We create train/validation/test splits of $47\,500/2\,500/10\,000$ images.
We preprocess the images by normalising all pixels per channel based on the training data.
We build the unsupervised contrastive learning dataset by considering
each of the $100$ label as a latent class,
using a block size of $2$ and a number of negative samples of~$4$
(see~\cref{sec:extensions} for the extended theory for block samples and more than one negative samples).

We also use \auslan~\citep{Kadous2002Thesis} dataset that contains $95$ labels, each one being a sign language's motion, and having $22$ dimensional features.
We split the dataset into $89\,775/12\,825/12\,825$ training/validation/test sets.
As pre-processing, we normalise feature vectors per dimension based on the training data.
The contrastive learning dataset then contains $95$ latent classes.
The block size and the number of negative samples are the same as \cifar{} setting.
More details are provided in~\cref{sec:auslan}.

\textbf{Neural networks architectures.}
For \cifar{} experiments,
we use a two hidden convolutional layers neural network (CNN).
The two hidden layers are convolutions (kernel size of $5$ and $64$ channels) with the ReLU activation function, followed by max-pooling (kernel size of $3$ and stride of $2$).
The final layer is a fully connected linear layer ($100$ neurons) without activation function.
For \auslan{} experiments,
we used a fully connected one hidden layer network with the ReLU activation function.
Both hidden and last layers have $50$ neurons.
More architecture details are given in \cref{sec:appendix_networks}.

\textbf{PAC-Bayes bound optimisation.}
We learn the network parameters by minimising the bound given by~\cref{theorem:iid-PAC-Bayesian-contrastive-bound}, using the strategy proposed in~\cref{sec:iid-algorithm}.
We rely on the logistic loss given by~\cref{eq:logloss}.
We fix the following PAC-Bayes bound's parameters: $b=100, c=0.1,$ and $\delta=0.05$.
The prior variance is initialised at $e^{-8}$.
The prior mean parameters $\boldsymbol{\mu}_{\Pcal}$ coincide with the random initialisation of the gradient descent.

We repeat the optimisation procedure with different combinations of hyper-parameters.
Namely, the PAC-Bayes bound constant $\lambda$ is chosen in $\{\frac{10^a}{m} | a{=}1,2,\ldots,9 \}$ for \cifar, and in $\{\frac{10^a}{m} | a{=}0,1,\ldots,5 \}$ for \auslan.
We also consider as a hyper-parameter the choice of the gradient descent optimiser, here between \emph{RMSProp}~\citep{Tieleman2012} and \emph{Adam}~\citep{Kingma2015ICLR}.
The learning rate is in $\{ 10^{-3}, 10^{-4} \}$. In all cases, $500$ epochs are performed and the learning rate is divided by $10$ at the $375^{\mathrm{th}}$ epoch.
To select the final model among the ones given by all these hyper-parameter combinations, we experiment three parameter selection criteria based on approaches described in \cref{sec:parameter-selection}, as detailed below. \\[1mm]
-- \emph{Stochastic validation} (\texttt{s-valid}).
This metric is obtained by randomly sampling $10$ set of network parameters according to the learnt posterior $\Qcal$, and averaging the corresponding empirical contrastive loss values computed on validation data.
The same procedure is used to perform \emph{early-stopping} during optimisation (we stop the learning process when the loss stops decreasing for $20$ consecutive epochs).\\[1mm]
-- \emph{Deterministic validation} (\texttt{det-valid}).
This metric corresponds to the empirical contrastive loss values computed on validation data of the deterministic network~$\fbf^*$, which corresponds to the mean parameters of the posterior (\ie, the \emph{maximum a posteriori} network given by~$\Qcal$).
Early stopping is performed in the same way as for \texttt{s-valid}.\\[1mm]
-- \emph{PAC-Bayes bound}  (\texttt{PB}).
The bound values of the learnt posterior $\Qcal$ are computed by using~\cref{eq:parameter-selection-criterion}.
Note that since this method does not require validation data, we perform optimisation over the union of the validation data and the training data.
We do not perform early stopping since the optimised objective function is directly the parameter selection metric.

\textbf{Benchmark methods.} We compare our results with two benchmarks, described below (more details are provided in \cref{sec:appendix_benchmarks})\\[1mm]
-- \emph{Prior contrastive unsupervised learning} \citep{Arora2019ICML}.
Following the original work, we minimise the empirical contrastive loss $\Lunhat(\fbf)$.
Hyper-parameter selection is performed on the validation dataset as for \texttt{s-valid} and \texttt{det-valid} described above.\\[1mm]
-- \emph{Supervised learning}  (\texttt{supervised}).
We also train the neural network in a supervised way, using the label information; Following the experiment of \citet{Arora2019ICML}, we add a prediction linear layer to our architectures (with $100$ output neurons for \cifar, and $95$ output neurons for \auslan{}), and minimise the multi-class logistic loss function
$$\ell_{\rm log} (\vbf) {\eqdef} \log_2(1+\textstyle\sum_{i=1}^{|Y|} e^{-v_i})\,.$$
Once done, we drop the prediction layer.
Then, we use the remaining network to extract feature representation.
\subsection{EXPERIMENTAL RESULTS}
\label{sec:experimental-results}
\begin{table*}[ht]
    \centering
    \caption{
        Supervised tasks results.
        \supervised{} was trained on the labelled training data,
        the others were trained on the contrastive training data.
        For \supervised, \citet{Arora2019ICML}, \texttt{s-valid}, and \texttt{det-valid},
        hyper-parameters were selected by using the validation loss.
        \texttt{PB} hyper-parameters were selected by the PAC-Bayes bound.
        The best scores are in bold among contrastive learning algorithms.
    }
    \vskip 0.1in
    {\small
        \begin{tabular}{@{}rrrrrrrrrrrrrrrrr@{}}
        \toprule
        & & & & & & & & \multicolumn{8}{c}{PAC-Bayes based methods} \\
        \cmidrule{9-16}
        & & \multicolumn{2}{c}{\supervised} & \phantom{a} & \multicolumn{2}{c}{\citet{Arora2019ICML}} & \phantom{a} & \multicolumn{2}{c}{\texttt{s-valid}} & \phantom{a} &
        \multicolumn{2}{c}{\texttt{det-valid}} & \phantom{a} & \multicolumn{2}{c}{\texttt{PB}} \\
        \cmidrule{3-4} \cmidrule{6-7} \cmidrule{9-10} \cmidrule{12-13} \cmidrule{15-16}
        &              & $\mu$ & $\mu$-$5$ && $\mu$ & $\mu$-$5$ && $\mu$  & $\mu$-$5$ && $\mu$   & $\mu$-$5$ && $\mu$ & $\mu$-$5$ \\ \midrule
        \multicolumn{2}{c}{\cifar}
        \\
        & \avgtwo  & $91.4$ & $87.5$ && $89.4$ & $85.6$ && $87.7$ & $83.9$ && $\mathbf{90.0}$ & $\mathbf{87.2}$ && $75.4$ & $70.8$ \\
        & \topone  & $25.3$ & $16.8$ && $\mathbf{22.5}$ & $15.6$ && $17.3$ & $12.7$ && $21.4$ & $\mathbf{16.0}$ && $6.9$ & $5.4$ \\
        & \topfive & $57.8$ & $46.0$ && $52.9$ & $42.6$ && $46.9$ & $38.3$ && $\mathbf{54.0}$ & $\mathbf{45.2}$ && $23.4$ & $19.4$ \\
        \midrule
        \multicolumn{2}{c}{\auslan } \\
        & \avgtwo  & $80.2$ & $75.1$ && $\mathbf{85.6}$ & $\mathbf{83.3}$ && $85.3$ & $82.7$ && $85.3$ & $82.9$ && $82.6$ & $79.1$ \\
        & \topone  & $12.0$ & $ 7.1$ && $\mathbf{38.0}$ & $\mathbf{24.9}$ && $36.1$ & $23.7$ && $37.1$ & $24.7$ && $23.2$ & $14.8$ \\
        & \topfive & $35.7$ & $24.1$ && $\mathbf{56.7}$ & $48.2$ && $56.2$ & $47.7$ && $56.5$ & $\mathbf{49.1}$ && $50.6$ & $38.4$ \\
        \bottomrule
        \end{tabular}
    }
    \label{tb:classification-result}
\end{table*}
\paragraph{Supervised classification.}
\cref{tb:classification-result} contains supervised accuracies obtained from the representation learnt with the two benchmark methods, as well as with our three parameter selection strategies on the PAC-Bayes learning algorithms.
For each method, two types of supervised predictor are used: $\mu$ and $\mu$-$5$ \citep[as in][]{Arora2019ICML}.\footnote{Our neural network architecture on \cifar{} differs from the one used in~\citet{Arora2019ICML}.
Their model is based on the deeper network \texttt{VGG-16}~\citep{Simonyan2015ICLR}, which explains why our accuracies are lower than the one reported in~\citet{Arora2019ICML}.}
The $\mu$ classifier is obtained  $\boldsymbol{\mu}_c$ that was the average vector of feature vectors $\hat{\fbf}_\wbf$ mapped from training data per supervised label,
and $\mu$-$5$ classifier had $\mubf_c$ that was average of $5$ random training samples feature vectors.
For $\mu$-$5$, we used averaged evaluation scores over $5$ times samplings on each experiment.

For the two datasets, we report three accuracies on the testing set, described below.
Values are calculated by averaging over three repetitions of the whole experiments using different random seeds.\\[1mm]
-- \emph{predictors-2 accuracy} (\avgtwo).
This is the empirical counterpart of \cref{eq:Rsupmuf}, \ie, given a test dataset $T\eqdef\{(\zbf_i, c_i)\}_{i=1}^{|T|}$ where $c_i \in \Ccal$ is a latent class, we define~$\avgtwo(\fbf_\wbf)\eqdef 1-\Rsuphat^\mu(\fbf_\wbf)$, given
\begin{align*}
    \Rsuphat^\mu(\fbf_\wbf) \eqdef \frac{C(C{-}1)}{2}
    \!\!\sum_{1\leq c^+ < c^ \leq C }\!\!
    \Rhat_{T_\cpm} (\hat g_\cpm \circ \hat \fbf_\wbf)\,,
\end{align*}
where $C$ is the number of latent classes (\eg, $C{=}100$ for \cifar{} dataset),
$\hat \fbf_\wbf$ is a feature map learnt from the training data,
$\hat g_{\cpm}(\hat \xbf) \eqdef (\hat\wbf_\cp - \hat\wbf_\cm) \cdot \hat \xbf$ is the predictor based on the centre of mass $\hat\wbf_\cp, \hat\wbf_\cm$ of the training data mapped features of classes $\cp, \cm$,
and $\Rhat_{T_\cpm}$ is the supervised risk on the dataset $T_\cpm \eqdef \{(\xbf,1)| (\xbf,c^+){\in} T\} \cup \{(\xbf,-1)| (\xbf,c^-){\in} T\} $:
\begin{align*}
    \Rhat_{T_\cpm}(\hat g_\cpm \circ \hat \fbf_\wbf)
    \eqdef \frac{1}{|T_\cpm|}
    \sum_{(\xbf, y)\in T_\cpm } r\big(\hat g_\cpm (\hat \fbf_\wbf(\xbf)), y \big)\,.
\end{align*}
-- \emph{Top-1 accuracy} (\topone{}).
This is the accuracy on the multi-class labelled test data $T$.
We predicted the label $\hat{y}_i = \argmax_y \boldsymbol{\mu}_y \cdot \fbf_\wbf(\xbf_i)$ on the test data.
Therefore,
$$\topone(\fbf_\wbf) \eqdef \frac{1}{|T|} \sum_{i=1}^{|T|} \onebf [y_i = \hat{y}_i]\,.$$
-- \emph{Top-5 accuracy} (\topfive{}).
For each test instance $(\xbf_i, y_i)\in T$, let $\hat{Y}_i$ be the set of  $5$ labels having the highest inner products $\boldsymbol{\mu}_y \cdot \fbf(\xbf_i)$. Then,
$$\topfive(\fbf_\wbf) \eqdef \frac{1}{|T|} \sum_{i=1}^{|T|} \onebf [y_i \in \hat{Y}_i]\,.$$
Note that the \topone{} and \topfive{} metrics are not supported by theoretical results, in the present paper or the work of \citet{Arora2019ICML}.
Nevertheless, we report those as an empirical hint of how representations are learnt by our contrastive unsupervised representation learning algorithm.

We observe that \texttt{det-valid} algorithm achieves competitive results with the ones of the CURL algorithm studied by \citet{Arora2019ICML}.
\begin{table}[ht]
    \centering
    \caption{
        Contrastive unsupervised PAC-Bayes bounds of the models used in \cref{tb:classification-result}.
    }
    \vskip 0.1in
    {\small
        \begin{tabular}{ccccc}\toprule
            & & \texttt{s-valid} & \texttt{det-valid} & \texttt{PB} \\
            \midrule
            \multicolumn{2}{l}{\cifar} \\
            & $\Runhat(\fbf^*)$ & $0.146$ & $0.131$ & $0.308$ \\
            & $\Run(\fbf^*)$ & $0.185$ & $0.167$ & $0.315$ \\
            & $\Runhat(\Qcal)$ & $0.172$ & $0.170$ & $0.323$ \\
            & $\Run(\Qcal)$ & $0.203$ & $0.197$ & $0.327$ \\
            & Bound & $0.733$ & $0.718$ & $0.437$ \\
            & $\KL$ & $32\,756$ & $30\,894$ & $1\,333$ \\
            & $\lambda\times m$ & $10^5$ & $10^5$ & $10^4$ \\
            & $\widehat{\lambda}\times m$ & $122\,781$ & $119\,687$ & $24\,295$ \\
            \midrule
            \multicolumn{2}{l}{\auslan} \\
            & $\Runhat(\fbf^*)$ & $0.193$ & $0.190$ & $0.263$ \\
            & $\Run(\fbf^*)$ & $0.182$ & $0.182$ & $0.216$ \\
            & $\Runhat(\Qcal)$ & $0.199$ & $0.195$ & $0.267$ \\
            & $\Run(\Qcal)$ & $0.186$ & $0.185$ & $0.220$ \\
            & Bound & $0.419$ & $0.417$ & $0.361$ \\
            & $\KL$ & $9\,769$ & $10\,018$ & $2\,054$ \\
            & $\lambda\times m$ & $10^5$ & $10^5$ & $10^4$ \\
            & $\widehat{\lambda}\times m$ & $95\,683$ & $97\,379$ & $45\,198$ \\
            \bottomrule
        \end{tabular}
    }
    \label{tb:pac-bayes-bounds}
\end{table}

\textbf{PAC-Bayesian generalisation bounds.}
\cref{tb:pac-bayes-bounds} shows the PAC-Bayes bound values obtained from~\cref{eq:parameter-selection-criterion}.
The bounds were calculated by using the same models used in~\cref{tb:classification-result}.
We also reported a training risk $\Runhat(\fbf^*)$ and test risk $\Run(\fbf^*)$ that we calculated by using only the mean parameter of the posterior as for neural network's weight.
The rows of $\widehat{\lambda}$ indicated the optimised $\lambda$ values that minimised~\cref{eq:parameter-selection-criterion}, and thus that correspond to the reported PAC-Bayes bounds.
Let us stress that all reported bounds values are non-vacuous.

The generalisation bounds obtained with the \texttt{PB} parameter selection criterion are naturally the tightest.
For this method, the gap between the empirical risk $\Runhat(\Qcal)$ and the test risk $\Runhat(\Qcal)$ is remarkably consistently small. This highlights that the PAC-Bayesian bound minimisation is not prone to overfitting.
On the downside, this behaviour seems to promote ``conservative'' solutions, which in turns gives lower supervised accuracy compared to methods relying on a validation set (see \cref{tb:classification-result}).

\section{CONCLUSION}
We extended the framework introduced by \citet{Arora2019ICML}, by adopting a PAC-Bayes approach to contrastive unsupervised representation learning.
This allows in particular to (i) derive new algorithms, by minimising the bounds (ii) remove the iid assumption.
While supported by novel generalisation bounds, our approach is also validated on numerical experiments are the bound yields non-trivial (non-vacuous) values.

\subsubsection*{Acknowledgements}

We thank Mikhail Khodak and Nikunj Saunshi for sharing their experimental setting and the reviewers for their fruitful comments,
and Louis Pujol, Ikko Yamane, and Han Bao for helpful discussions.
This work was supported by the French Project APRIORI ANR-18-CE23-0015 and BEAGLE ANR-18-CE40-0016-01.
KN is supported by JSPS KAKENHI Grant Number 18J20470.
PG is supported by the Canada CIFAR AI Chair Program.

\bibliography{reference}
\bibliographystyle{plainnat}
\newpage
\onecolumn
\appendix

\section{EXTENDED PAC-BAYES BOUNDS}
\label{sec:extensions}
\citet{Arora2019ICML} show two extended generalisation error bounds based on~\cref{theorem:Arora-bound}.
We also show each PAC-Bayesian counterpart of their extended bounds for~\cref{theorem:iid-PAC-Bayesian-contrastive-bound}.
In addition, we show PAC-Bayesian analysis of a general supervised classifier instead of the mean classifier.
\subsection{BLOCK BOUND}
\label{sec:block-bound-with-iid}
The first extension is to use block pairs for positive and negative samples to make the bound tighter.
We also derive a tighter PAC-Bayes bound in the same setting.

Let $b$ be the size of blocks.
We change the data generation process;
Given $(c^+, c^-) \sim \rho^{2}$, we sample $(\xbf, \{\xbf_j^+\}_{j=1}^{b}) \sim \Dcal_{c^+}^{b+1}$ and $\{\xbf_j^-\}_{j=1}^b \sim \Dcal_{c^-}^b$.
Given block pairs, unsupervised block loss is defined as
\begin{align}
    \Lun^{block}(\fbf) = \bbE \left \{
        \ell \left[ f(\xbf) \cdot \left(
            \frac{\sum_{i=1}^b \fbf(\xbf_i^+)}{b}
            -
            \frac{\sum_{i=1}^b \fbf(\xbf_i^-)}{b}
            \right) \right]
    \right\}.
\end{align}

This block loss $\Lun^{block}(\fbf)$ lower bounds $\Lun(\fbf)$~\citep[Proposition 6.2]{Arora2019ICML}: $\forall \fbf \in \Fcal, \Lun^{block}(\fbf) \leq \Lun(\fbf)$.
Based on this lower bound, when we define $\Lun^{block}(\Qcal) = \bbE_{\fbf \sim \Qcal} \Lun^{block}(\fbf)$,
we obtain the following lower bound of the unsupervised risk $\Lun(\Qcal)$ for all $\Qcal$ over $\Fcal$ by taking the expected value according to $\Qcal$,
\begin{align*}
    \Lun^{block}(\Qcal) \leq \Lun(\Qcal).
\end{align*}
Therefore we derive the tighter block bound by combining the previous lower bound and~\cref{theorem:iid-PAC-Bayesian-contrastive-bound}.
\begin{proposition}
    $\forall \Qcal$ over $\Fcal$,
    \begin{align}
        \Lsup^{\mu}(\Qcal)
        & \leq
        \frac{1}{1 - \tau}
        \left(
            B_\ell
            \frac{
                    1 - \exp \left(
                            - \frac{\lambda}{B_\ell} \Lunhat^{block}(\Qcal)
                            - \frac{\KL(\Qcal \| \Pcal) + \ln\frac{1}{\delta}}{m}
                        \right)
                }{
                    1 - \exp(-\lambda)
                }
            - \tau
        \right).
    \end{align}
\end{proposition}
\subsection{$k$-NEGATIVE SAMPLES BOUND}
\label{sec:k-neg-bound}
The second extension is to use $k$ negative samples in their framework as a general setting.
Following \citet{Arora2019ICML}, we consider the data generation process with $k$ negative samples per each pair.
Let $\Ucal$ be the process that generates an unlabelled sample $\zbf=(\xbf, \xbf^+, \{\xbf^-_i \}_{i=1}^k)$ according to the following scheme:

1. Draw $k+1$ latent classes $(c^+, \{c^-_i\}_{i=1}^k) \sim \rho^{k+1}$\,;\\
2. Draw two similar samples $(\xbf, \xbf^+) \sim (\Dcal_{c^{+}})^2$\,;\\
3. Draw $k$ negative samples $\{ \xbf^-_i \sim \Dcal_{c^-_i} \mid i = 1, \ldots, k \} $\,.

We extend loss functions for a vector of size $k$.
We use two convex loss functions:
\begin{align}
\label{eq:k-logloss}
    \ell_{\rm log} (\vbf) \eqdef& \log_2 (1 + \sum_{i=1}^k e^{-v_i} ) \,, & \mbox{\it(logistic loss)}\\
\label{eq:k-hingeloss}
    \ell_{\rm hinge} (\vbf) \eqdef& \max[0, 1 + \max_i (-v_i) ] \,, & \mbox{\it (hinge loss)}
\end{align}
Then we define unsupervised contrastive loss and empirical contrastive loss with $k$ negative samples;
\begin{align}
    \Lun(\fbf) \eqdef& \bbE_{\zbf \sim \Ucal }
    \ell \left(
        \left\{
            \fbf(\xbf) \cdot \big[ \fbf(\xbf^+) - \fbf(\xbf^-_{i}) \big]
        \right\}_{i=1}^k
    \right), \\
    \Lunhat(\fbf) \eqdef&
    \frac{1}{m} \sum_{i=1}^m
    \ell \left(
        \left\{
            \fbf(\xbf_i) \cdot \big[ \fbf(\xbf_i^+) - \fbf(\xbf^-_{ij}) \big]
        \right\}_{j=1}^k
    \right).
\end{align}
We analyse a mean classifier as with $k=1$ scenario.
Let $\Tcal$ be the set of supervised classes whose size is $k+1$,
let $\Dcal$ be the distribution over $\Tcal$,
and let $\Dcal_{\Tcal}$ be the distribution over class in $\Tcal$.
The supervised average loss of mean classifier with $k$ negative samples is defined as
\begin{align}
    \Lsup^{\mu}(\fbf) = \bbE_{\Tcal \sim \Dcal} \Lsup^{\mu}(\Tcal, \fbf) = \bbE_{\Tcal \sim \Dcal} \bbE_{ c \sim \Dcal_{\Tcal} } \bbE_{\xbf \sim \Dcal_c} [ \ell ( \{ \fbf(\xbf) \cdot (\mubf_{c} - \mubf_{c'})  \}_{c' \neq c} ) ].
\end{align}

To introduce the counterpart of~\cref{lemma:sup-bound} for $k$ negative samples,
we introduce notations related to the extended class collision.
Let $I^+(c_1^-, \ldots , c_k^-) = \{ i \in [1, \ldots, k] \mid c^-_{j} = c^+ \}$ be a set of negative sample indices such that $c_j$ is the same to $c^+$.
Let $\tau_k = P(I^+ \neq \phi)$ be the class collision probability,
and let $Q$ be a distinct latent class set of $c^+, c^-_1, \ldots, c^-_k$ sampled from $\rho^{k+1}$.

The following \cref{lemma:sup-bound-k} shows the upper bound of supervised average loss with $k \geq 1$ by the unsupervised contrastive loss.
\begin{lemma}\citep[Eq. 26]{Arora2019ICML},
    \label{lemma:sup-bound-k}
    $\forall \fbf \in \Fcal$,\footnote{In the original paper from \cite{Arora2019ICML}, it is shown for $\widehat{\fbf}$, but actually it is valid $\forall \fbf \in \Fcal$.}
    \begin{align}
        (1 - \tau_k) \bbE_{\Tcal \sim \Dcal}
        \frac{p^+_{\min}(\Tcal)}{p_{\max}(\Tcal)}
        \Lsup^{\mu}(\Tcal, \fbf)
        \leq \Lun(\fbf)
        - \tau_k \bbE_{c^+, \{c^-_i \}_{i=1}^k \sim \rho^{k+1}} [\ell ( \mathbf{0}_{|I^+|} ) \mid I^+ \neq \phi].
    \end{align}
    where $\mathbf{0}_{|I^+|}$ is zero vector of size $|I^+|$,
    $p_{\max}(\Tcal) = \max_c \Dcal_{\Tcal}(c)$,
    and \\
    $p^+_{\min}(\Tcal) = \min_{c \in \Tcal} p_{c^+, \{c^-_i\}_{i=1}^k \sim \rho^{k+1}} (c^+ = c \mid \Tcal = Q, I = \phi )$.
\end{lemma}
Let us denote $\Qcal$-weighted loss functions of contrastive learning with $k$ negative samples:
\begin{align}
    \Lun(\Qcal) \eqdef& \bbE_{\fbf \sim \Qcal} \Lun(\fbf), \\
    \Lunhat(\Qcal) \eqdef& \bbE_{\fbf \sim \Qcal} \Lunhat(\fbf), \\
    \Lsup^{\mu}(\Tcal, \Qcal) \eqdef& \bbE_{\fbf \sim \Qcal} \Lsup^{\mu}(\Tcal, \fbf).
\end{align}
We derive the following~\cref{theorem:theorem:PAC-Bayesian-contrastive-bound-k} based on~\cref{lemma:sup-bound-k} to extend ~\cref{theorem:iid-PAC-Bayesian-contrastive-bound} for $k \geq 1$.
\begin{theorem}
    \label{theorem:theorem:PAC-Bayesian-contrastive-bound-k}
    Let $B \in \Rbb_+$ such that $\|\fbf(\cdot)\| \leq B$ for all $\fbf\in\Fcal$.
    Given $k \in \Nbb, \lambda > 0$ and a prior $\Pcal$ over $\Fcal$,
    with probability at least $1-\delta$ over training samples $U \sim \Ucal^m$, $\forall \Qcal$ over $\Fcal$,
    \begin{multline*}
        (1 - \tau_k) \bbE_{\Tcal \sim \Dcal}
        \frac{p^+_{\min}(\Tcal)}{p_{\max}(\Tcal)}
        \Lsup^{\mu}(\Tcal, \Qcal)
        \leq
        \\
            B_\ell \frac{
                1 - \exp \left(
                    - \frac{\lambda}{B_\ell} \Lunhat(\Qcal)
                    - \frac{\KL(\Qcal \| \Pcal) + \ln\frac{1}{\delta}}{m}
                \right)
            }{
                1 - \exp \left(-\lambda \right)
            }
        - \tau_k \bbE_{c^+, \{c^-_i \}_{i=1}^k \sim \rho^{k+1}} [\ell ( \{ 0 \}_{|I^+|} )  \mid I^+ \neq \phi].
    \end{multline*}
    with $B_\ell{\eqdef} \log_2 (1 + k e^{2B^2})$ for the logistic loss,
    or $B_\ell{\eqdef} 1 + 2B^2$ for the hinge loss.
\end{theorem}
\begin{proof}
    We follow similar steps to the proof of \cref{theorem:iid-PAC-Bayesian-contrastive-bound}.
    Since $\|\fbf(\cdot)\| \leq B$, we have $\forall \xbf,\xbf^+,\xbf^-\in X^3$:
    $$-2 B^2 \leq  \fbf(\xbf) \cdot [ \fbf(\xbf^+)-\fbf(\xbf^-) ] \leq 2 B^2\,. $$
    Given the number of negative samples $k$,
    from the loss functions' definition, we can obtain the lower bound and upper bound explicitly.
    \begin{align}
        \log_2 (1 + k e^{-2B^2} ) &\leq \ell_{\rm log} (\vbf) \leq \log_2 (1 + k e^{2B^2} ) \,,
        \\
        0 &\leq \ell_{\rm hinge} (\vbf) \leq 1 + 2B^2 \,,
    \end{align}
    Thus $B_{\ell_{\rm log}} \eqdef \log_2 (1 + k e^{2B^2})$ and $B_{\ell_{\rm hinge}} \eqdef 1 + 2B^2$.
    Therefore we can bound the $\Lun(\Qcal)$ by using the same inequality in the proof of \cref{theorem:iid-PAC-Bayesian-contrastive-bound}:
    With probability at least $1-\delta$,
    \begin{align*}
        \frac{1}{B_\ell} \Lun(\Qcal) \leq
            \frac{
                1 {-} \exp \left(
                    - \lambda\frac{1}{B_\ell}\Lunhat(\Qcal)
                    - \frac{\KL(\Qcal \| \Pcal) + \ln\frac{1}{\delta}}{m}
                \right)
            }{
                1 - \exp \left(-\lambda \right)
            }\,.
    \end{align*}
    Also since \cref{lemma:sup-bound-k} is true for all $\fbf \in \Fcal$,
    we take expected value according to $\Qcal$;
    \begin{align}
        (1 - \tau_k) \bbE_{\Tcal \sim \Dcal}
        \frac{p^+_{\min}(\Tcal)}{p_{\max}(\Tcal)}
        \Lsup^{\mu}(\Tcal, \Qcal)
        & \leq \Lun(\Qcal)
        - \tau_k \bbE_{c^+, \{c^-_i \}_{i=1}^k \sim \rho^{k+1}} [\ell ( \{ 0 \}_{|I^+|} )  \mid I^+ \neq \phi].
    \end{align}
    The result is obtained by replacing $\Lun(\Qcal)$ in the above inequality by its bound in terms of $\Lunhat(\Qcal)$.
\end{proof}
\subsection{LOWER BOUND OF GENERAL CLASSIFIER}
\label{sec:general-analysis}
We give PAC-Bayesian analysis of a general classifier's lower bound by the similar way to~\citet[Theorem~4.5]{Arora2019ICML}.
More precisely, we derive a lower bound of \cref{theorem:iid-PAC-Bayesian-contrastive-bound} by using a general classifier instead of the mean classifier.

We introduce a posterior distribution, $\Qcal_{\mathrm{sup}}$, over hypothesis class of predictors $\Gcal = \{g \in \Rbb^d \to \Rbb\}$.
Given $\Qcal$ trained on unsupervised data, we define the PAC-Bayesian supervised loss as
\begin{align*}
    L_{\mathrm{sup}}(\Qcal_{\mathrm{sup}}, \Qcal) \eqdef
    \bbE_{g \sim \Qcal_{\mathrm{sup}}}
    \bbE_{\fbf \sim \Qcal}
    \Lsup(g \circ \fbf).
\end{align*}
The optimal posterior is denoted $\widehat{\Qcal}_{\mathrm{sup}} = \argmin_{\Qcal_{\mathrm{sup}}} L_{\mathrm{sup}}(\Qcal_{\mathrm{sup}}, \Qcal)$.
Therefore we derive a lower bound of~\cref{theorem:iid-PAC-Bayesian-contrastive-bound}:
\begin{align}
    \Lsup(\widehat{\Qcal}_{\mathrm{sup}}, \Qcal) \leq \Lsup^{\mu}(\Qcal).
\end{align}
Therefore general classifier's loss is at least as good as the mean classifier's loss.
\section{CONTRASTIVE ZERO-ONE RISK WITH $k$-NEGATIVE SAMPLES}
\label{sec:contrastive-k-zero-one-risk}
We extend the zero-one risk to $k$ negative sampling setting;
Let $\zbf = (\xbf, \xbf^+, \xbf^-_1, \ldots, \xbf^-_k)$, then
\begin{align}
    r_k(\zbf) = \frac1k \sum_{i=1}^k
    r(\fbf(\xbf^+) - \fbf(\xbf^-_i), \fbf(\xbf)).
\end{align}
We use this zero-one risk to compute $\Rhat(\Qcal)$ used in~\cref{eq:parameter-selection-criterion}.

\section{FULL EXPRESSION OF $\chi^2$-DIVERGENCE}
\label{sec:full-chi-square-divergence}
From \citet[Eq. 8.52]{Bock2000Book-chapter},
$\chi^2$-square divergence between our posterior and prior has the closed-form:
\begin{multline}
    \chi^2(\Qcal \| \Pcal ) =
    \frac{
        \frac{1}{\sigma^{2N}_\Pcal} \left| \diag(\sigmabf{}_\Qcal^2) \right|
    }{
        \sqrt{ \left| \frac{2}{\sigma^2_\Pcal} \diag(\sigmabf{}_\Qcal^2) - I \right|}
    }
    \\
    \exp \left\{
        \frac12\left(
            \left\|
                \frac{2}{\sigma^2_\Pcal} \mubf_\Pcal - \diag(\sigmabf{}_\Qcal^2)^{-1} \mubf_\Qcal
            \right\|^2_{
                \left[ \frac{2}{\sigma^2_\Pcal} I - \diag(\sigmabf{}_\Qcal^2)^{-1}) \right] ^{-1}
            }
            + \| \mubf_\Qcal \|^2_{\diag(\sigmabf_{\Qcal}^2)^{-1}}
            - \frac{2}{\sigma^2_\Pcal} \| \mubf_\Pcal \|^2
        \right)
    \right\} - 1.
\end{multline}
where $\| \xbf \|_{A}$ is the Mahalanobis distance $\sqrt{ \xbf^\top A \xbf }$.
Note $\frac{2}{\sigma^2_\Pcal} I - \diag(\sigmabf{}_\Qcal^2)^{-1}$ is positive definite.\footnote{To hold positive definite of the matrix, we replace each posterior variance parameter $\sigmabf{}^2_{\Qcal, i}$ with $\frac12 \sigma_{\Pcal}^2$ if $\sigmabf{}^2_{\Qcal, i} < \frac12 \sigma_{\Pcal}^2$ during optimisation.}

\section{EXPERIMENTAL DETAILS}
\subsection{DATASETS}
\label{sec:auslan}
\paragraph{\auslan{} dataset.}
We used \auslan{} time-series dataset instead of \texttt{Wiki-3029} used in \citet{Arora2019ICML}, which contains $3\,029$ classes' sentences sampled from Wikipedia.
This is because \citet{Arora2019ICML} used recurrent neural networks on this dataset,
but PAC-Bayes theory with recurrent neural networks on word sequences dataset is not trivial due to its time-dependent predictor and data sparsity,
so it is not out of scope in this paper.
Therefore we selected \auslan{} dataset as a simpler and similar dataset.

\auslan{} originally contains $27$ time-series samples per class.
Each sample has different lengths, whose the maximum is $136$ and the minimum is $45$,
and each time step is represented by a feature vector whose dimensionality is $22$.
We treated each feature vector as each input sample in our experiment.
In addition, we sample the first $45$ time steps from each original time series to unify the number of samples per class.
We separated original $27$ times-series into $24/3$ training/test sets.
Then we selected $3$ time-series of training dataset per class randomly as a validation set for each random seed.
Thus, we obtained $89\,775/12\,825/12\,825$ training/validation/test datasets.
We used these datasets as supervised datasets.
We created contrastive datasets in the same way to the \cifar's experiment.
\subsection{NETWORK ARCHITECTURES AND INITIALISATION PARAMETERS}
\label{sec:appendix_networks}
\paragraph{\cifar{} experiments.}
For all convolution layers, the number of channels was $64$, the kernel size was $5$, the stride of the convolution was $1$, zero-padding was $1$, and the dilation was $1$.
The convolutional layers' parameters were initialised as zero-mean truncated Gaussian distribution whose $\sigma$ was $0.1$.
For all max-pooling layers, the kernel size was $3$, the stride of the window was $2$, and the dilation was $1$.
For the linear layer, the number of units was $100$.
The linear layers' parameters were also initialised as zero-mean truncated Gaussian distribution whose $\sigma$ was $1/800$.
For all convolutional layers and linear layers, biases were initialised as $0$.
\paragraph{\auslan{} experiment.}
We used a fully connected one hidden layer's network with ReLU activation function.
Both hidden and last layer have $50$ neurons.
The hidden layers' parameters were initialised as zero-mean truncated Gaussian distribution whose $\sigma$ was $1/11$,
and the output layer's parameters were initialised as zero-mean truncated Gaussian distribution whose $\sigma$ was $1/25$.
\subsection{BENCHMARK METHODS}
\label{sec:appendix_benchmarks}
\paragraph{Comparison with \citet{Arora2019ICML}.}
We optimised the model by using a stochastic gradient descent algorithm with $100$ mini-batches and $500$ epochs.
We searched the best learning rate in $\{ 10^{-3}, 10^{-4}\}$
and optimiser algorithm in stochastic gradient descent (SGD) with momentum $0.9$, \emph{RMSProp}, and \emph{Adam}.
We also performed early-stopping and updated the learning rate by the same as the PAC-Bayes setting.
\paragraph{Supervised learning.}
The additional linear layers' parameters were initialised as zero-mean truncated Gaussian distribution with $\sigma = 1/50$, and a bias was initialised as $0$.
The loss function was the multi-class logistic loss.
We did the same way to find the best hyper-parameters, learning rate and optimiser, and to perform early-stopping.
Optimisation methods and procedures were also the same as the non-PAC-Bayesian contrastive learning setting.
\section{NON-IID EXPERIMENTS}
\label{sec:non-iid-experiments}
We conduct experiments by using the algorithm in \cref{sec:non-iid-algorithm} on contrastive data without iid assumption.
\subsection{PARAMETER SELECTION}
For parameter selection with respect to optimiser and learning rate,
we can use the same strategies based on validation data: \texttt{s-valid} and \texttt{det-valid}, which are described in \cref{sec:parameter-selection}.

As a counterpart to the PAC-Bayes bound criterion~\eqref{eq:parameter-selection-criterion},
we select a trained model with best hyper-parameters such that it minimises the following PAC-Bayes bound:
\begin{align}
    \label{eq:parameter-selection-criterion-non-iid}
    \Runhat(\Qcal)
    +
    \pi j
    \sqrt{
        \frac{1}{24m \delta} ( 1 + 8T )
        \left(
            \chi^2(\Qcal \| \Pcal) + 1
        \right)
    }.
\end{align}
Note that we do not optimise the bound with respect to $\lambda$ like \cref{eq:parameter-selection-criterion} after minimising \cref{eq:non-iid-objective},
since there is no $\lambda$ parameter in \cref{eq:parameter-selection-criterion-non-iid}.
\subsection{DATASET}
\label{sec:non-iid-auslan}
We create \noniidauslan{} for our non-iid data experiments by modifying creation procedures of the \auslan{} dataset.
We make positive pairs such that adjacent samples in the original time-series are treated as similar samples.
Formally, we create positive pair $(\xbf_t, \{\xbf_{j}\}_{j=t+1}^{t+B} )$, $t=1, \dots (45-B)$ per original sample.
In these experiments, we used the block size $B = 2$, which also means $T=2$ in the non-iid objective.
Negative pairs are created in the same way to the \cifar's experiment.
As a result, we obtained $85\,785/12\,255/12\,255$ training/validation/test contrastive datasets.
Supervised datasets are exactly same as \auslan{} datasets.
\subsection{NETWORK ARCHITECTURES AND INITIALISATION PARAMETERS}
We use the same settings described in \cref{sec:appendix_networks} excepting that prior's variance is initialised at $e^{-5}$.
\subsection{OPTIMISATION}
Optimisers and their hyper-parameters are same as \auslan's PAC-Bayes setting.
\subsection{BENCHMARK METHODS}
There is no competitor for non-iid bound based algorithm because other algorithms are derived from the generalisation bounds requiring iid assumption.
As references, we report the performances of \citet{Arora2019ICML}'s algorithm and our CURL algorithms proposed in \cref{sec:iid-algorithm}.
Their hyper-parameters and networks are same as \auslan's experimental settings.

\subsection{RESULTS}
\cref{tb:non-iid-classification-result} reports classification performance on supervised data.
The classification performances do not perform well like iid results shown by \cref{tb:classification-result}.
We believe that $\chi^2$-divergence causes poor classification performance because its value rapidly increases when the posterior moves from the prior.

\begin{table*}[tbh]
    \centering
    \caption{
        Supervised tasks results on \noniidauslan.
        All methods were trained on the contrastive training data.
        For \citet{Arora2019ICML}, \texttt{s-valid}, and \texttt{det-valid},
        hyper-parameters were selected by using the validation loss.
        \texttt{PB} hyper-parameters were selected by the PAC-Bayes bounds.
    }
    \vskip 0.1in
    {\tiny
        \begin{tabular}{@{}rrrrrrrrrr|rrrrrrrrrrrr@{}}
        \toprule
        & \multicolumn{8}{c}{} & & \multicolumn{11}{c}{As Reference} \\
        \cmidrule{12-22}
        & \multicolumn{8}{c}{non-iid bound based Algorithms (\cref{sec:non-iid-algorithm}) } & & &  \multicolumn{2}{c}{} & & \multicolumn{8}{c}{iid bound based Algorithms (\cref{sec:iid-algorithm}) } \\
        \cmidrule{2-9}
        \cmidrule{15-22}
        & \multicolumn{2}{c}{\texttt{s-valid}} & &
        \multicolumn{2}{c}{\texttt{det-valid}} & & \multicolumn{2}{c}{\texttt{PB}}
        & & &
        \multicolumn{2}{c}{\citet{Arora2019ICML}} & & \multicolumn{2}{c}{\texttt{s-valid}} & &
        \multicolumn{2}{c}{\texttt{det-valid}} & \phantom{a} & \multicolumn{2}{c}{\texttt{PB}} \\
        \cmidrule{2-3} \cmidrule{5-6} \cmidrule{8-9}
        \cmidrule{12-13}
        \cmidrule{15-16} \cmidrule{18-19} \cmidrule{21-22}
                 & $\mu$ & $\mu$-$5$ && $\mu$ & $\mu$-$5$ && $\mu$ & $\mu$-$5$ & & & $\mu$ & $\mu$-$5$ && $\mu$  & $\mu$-$5$ && $\mu$ & $\mu$-$5$ && $\mu$ & $\mu$-$5$ \\ \midrule
        \avgtwo  & $69.2$ & $64.0$ && $69.7$ & $64.3$ && $69.4$ & $64.2$ &&& $75.2$ & $67.3$ && $74.6$ & $66.4$ && $74.8$ & $66.5$ && $72.3$ & $64.0$ \\
        \topone  & $ 6.8$ & $ 4.9$ && $ 6.9$ & $ 4.9$ && $ 6.9$ & $ 5.0$ &&& $20.5$ & $ 9.6$ && $19.6$ & $ 9.1$ && $21.0$ & $ 9.3$ && $14.4$ & $ 7.5$ \\
        \topfive & $22.0$ & $17.1$ && $22.1$ & $17.3$ && $22.5$ & $17.3$ &&& $41.7$ & $23.4$ && $40.8$ & $22.8$ && $41.7$ & $22.9$ && $34.9$ & $20.1$ \\
        \bottomrule
        \end{tabular}
    }
    \label{tb:non-iid-classification-result}
\end{table*}

\cref{tb:non-iid-auslan-pac-bayes-bounds} shows the PAC-Bayes bound values obtained from~\cref{eq:parameter-selection-criterion-non-iid}.
All bounds are vacuous, but the gap between the generalisation risk and the training risk tends to be small.
\begin{table*}[tbh]
    \centering
    \caption{
        Contrastive unsupervised PAC-Bayes bounds of the models used in \cref{tb:non-iid-classification-result}.
    }
    \vskip 0.1in
    {
        \begin{tabular}{cccc}
            \toprule
            & \texttt{s-valid}& \texttt{det-valid} & \texttt{PB} \\
            \midrule
            $\Runhat(\fbf^*)$ & $0.056$ & $  0.073$ & $0.033$ \\
            $\Run(\fbf^*)$    & $0.089$ & $  0.103$ & $0.074$ \\
            $\Runhat(\Qcal)$  & $0.058$ & $  0.065$ & $0.058$ \\
            $\Run(\Qcal)$     & $0.106$ & $  0.108$ & $0.097$ \\
            Bound             & $4.460$ & $140.949$ & $2.227$ \\
            $\chi^2$          & $0.012$ & $980.354$ & $0.052$ \\
            \bottomrule
        \end{tabular}
    }
    \label{tb:non-iid-auslan-pac-bayes-bounds}
\end{table*}
\end{document}